\documentclass[10pt,twocolumn,letterpaper]{article}

\usepackage[pagenumbers]{cvpr} %
\usepackage{xspace} %
\usepackage{booktabs}
\usepackage{multirow}
\usepackage{graphicx}
\usepackage{placeins}

\newcommand{\mname}{~{SPEAR-1}\xspace}
\newcommand{\mnamevlm}{~{SPEAR-VLM}\xspace}

\newcommand{\mhat}[1]{#1}

\definecolor{cvprblue}{rgb}{0.21,0.49,0.74}
\usepackage[pagebackref,breaklinks,colorlinks,allcolors=cvprblue]{hyperref}

\def\paperID{46344} %
\def\confName{CVPR}
\def\confYear{2026}

\title{\mname: Scaling Beyond Robot Demonstrations via 3D Understanding}

\author{
\begin{tabular}{cccc}
Nikolay Nikolov & Giuliano Albanese & Sombit Dey & Aleksandar Yanev \\
\multicolumn{4}{c}{Luc Van Gool \quad Jan-Nico Zaech \quad Danda Pani Paudel}
\end{tabular}
\\[0.5em]
{\normalsize \texttt{\{%
nikolay.nikolov,
giuliano.albanese,
sombit.dey,
aleksandar.yanev}}\\
{\normalsize \texttt{%
luc.vangool,
jan-nico.zaech,
danda.paudel\}@insait.ai}}
\\[0.75em]
{\normalsize INSAIT, Sofia University "St. Kliment Ohridski"}
}

\begin{document}

\twocolumn[{%
\renewcommand\twocolumn[1][]{#1}%
\maketitle

\begin{center}
    \includegraphics[width=\textwidth, trim={1cm 0cm 1cm 0cm}]{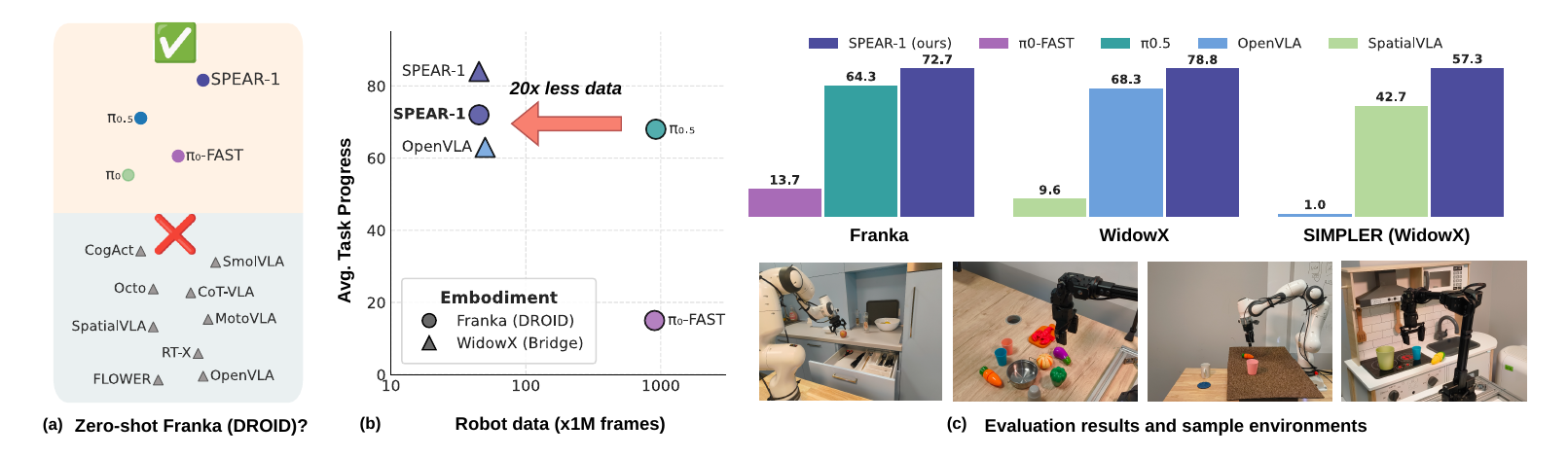}
    \captionof{figure}{
        (a) Most VLAs fail to show zero-shot performance on the challenging Franka (DROID) setup in unseen environments, without task or environment-specific fine-tuning.
        (b) \mname operates in this challenging setup, outperforms $\pi0$-FAST~\citep{pertsch2025fast} and matches $\pi_{0.5}$~\citep{intelligence2025pi_} on Franka (DROID) embodiment zero-shot in unseen environments while using 20$\times$ less robot demonstrations data.
        It also shows strong performance on WidowX (Bridge)
        (c) \mname evaluation results on different embodiments and in different environments.
    }
    \label{fig:teaser_fig}
\end{center}

}]

\begin{abstract}

Robotic Foundation Models (RFMs) hold great promise as generalist, end-to-end systems for robot control.
Yet their ability to generalize across new environments, tasks, and embodiments remains limited.
We argue that a major bottleneck lies in their foundations: most RFMs are built by fine-tuning internet-pretrained Vision-Language Models (VLMs).
However, these VLMs are trained on 2D image-language tasks and lack the 3D spatial reasoning inherently required for embodied control in the 3D world.
Bridging this gap directly with large-scale robotic data is costly and difficult to scale.
Instead, we propose to enrich easy-to-collect non-robotic image data with 3D annotations and enhance a pretrained VLM with 3D understanding capabilities.
Following this strategy, we train SPEAR-VLM, a 3D-aware VLM that infers object coordinates in 3D space from a single 2D image.
Building on SPEAR-VLM, we introduce our main contribution, ~\textbf{SPEAR-1}: a robotic foundation model that integrates grounded 3D perception with language-instructed embodied control.
Trained on $\sim$45M frames from 24 Open X-Embodiment datasets, SPEAR-1 outperforms or matches state-of-the-art models such as $\pi_0$-FAST and $\pi_{0.5}$, while it uses 20$\times$ fewer robot demonstrations.
This carefully-engineered training strategy unlocks new VLM capabilities and as a consequence boosts the reliability of embodied control beyond what is achievable with only robotic data.
We make our model weights and 3D-annotated datasets publicly available at \href{https://spear.insait.ai}{spear.insait.ai}.

\end{abstract}

\section{Introduction}
\label{sec:intro}

Vision-Language-Action (VLA) models have emerged as a promising paradigm for building generalist, end-to-end systems for robot control. Their success relies on two factors: (1) the strong visual-linguistic understanding inherited from internet-scale pretraining of the underlying Vision Language Model (VLM), which provides broad “common sense” knowledge, and (2) training on large, diverse datasets of robot demonstrations.

Despite this progress, the landscape of generalist VLA policies remains fragmented in terms of generalization -- across embodiments, environments, and tasks.
This becomes especially prominent in zero-shot performance in ~\emph{unseen} real-world environments with variations in camera positions and out-of-distribution backgrounds, such as the typical deployment scenarios of the Franka (DROID) setup~\citep{atreya2025roboarena}.
In contrast, as shown in \cref{fig:teaser_fig} (a), most existing VLAs (\eg OpenVLA~\citep{kim2024openvla}, CogAct~\citep{li2024cogact}, SpatialVLA~\citep{qu2025spatialvla}, MotoVLA~\citep{spiridonov2025generalist}) achieve high zero-shot performance\footnote{For more details on the definition of "zero-shot performance in unseen environments" see Appendix \ref{subsec:app-zero-shot-perf}} in “toy” environments with seen camera positions, but struggle with zero-shot performance in ~\emph{unseen} challenging Franka scenarios and depend on task- or environment-specific finetuning.
Recent efforts such as $\pi_0$ and $\pi_{0.5}$~\cite{intelligence2025pi_} push toward broader generalization, yet at the cost of closed large-scale robotic data.
We introduce \textbf{\mname}, which advances these desired generalization capabilities while being substantially more data-efficient. Quantitatively (see \cref{fig:teaser_fig} (b)), \textbf{\mname} outperforms $\pi_0$-FAST~\cite{pertsch2025fast} and matches $\pi_{0.5}$ on multiple robot embodiments using $20\times$ fewer demonstrations, which is especially important given the high cost and logistical difficulty of collecting real-world robotic data.

We achieve this efficiency by introducing explicit 3D awareness into the vision-language backbone before any robot training.
The model incorporates a pretrained depth encoder and is optimized on 3D-aware vision-language tasks such as distance estimation and 3D bounding box prediction, embedding control-relevant spatial reasoning directly into its representations.
Achieving such integration is non-trivial: aligning 3D geometric cues with high-level linguistic and visual features requires detailed multimodal dataset annotations and precise cross-modal calibration, as naive fusion often degrades both semantic understanding and spatial accuracy.
In contrast, existing VLAs rely on 2D VLMs that excel at semantic perception, but lack geometric understanding, forcing them to learn 3D structure implicitly from large-scale robot demonstrations.
This dependence on costly and embodiment-specific data limits scalability and generalization across environments, underscoring the difficulty and significance of \mname's design.

\begin{figure}[t]
    \centering
    \includegraphics[trim={6mm 0 6mm 0},clip, width=\linewidth]{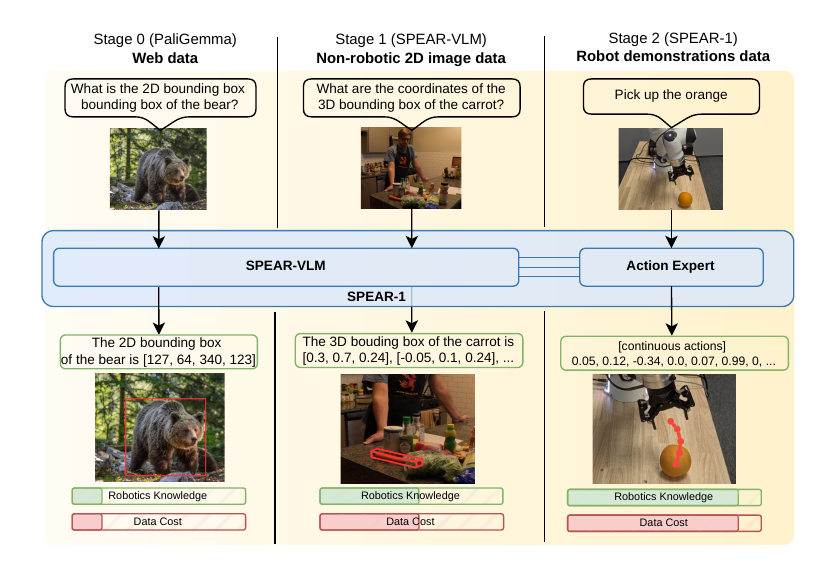}    
    \caption{
        \textbf{\mname stages of training.}
        \textbf{Stage 0}: General VLM pretraining on web scale data, \eg PaliGemma.
        \textbf{Stage 1}: Integrate a mono depth vision encoder to build \textbf{\mnamevlm} and train it on embodied-inspired VQA tasks, \eg 3D bounding box or object-to-object distance estimation. We use 2D images from non-robotic data, enriched with 3D annotations. \textbf{Stage 2}: Add an \textit{action expert} to train \textbf{\mname} on robot demonstration data, \eg OpenX~\citep{open_x_embodiment_rt_x_2023}.
        Each stage boosts the model's robotics-relevant knowledge and capabilities, but the abundance and diversity of data decreases.
    }
    \label{fig:teaser}
    \vspace{-10pt}
\end{figure}

In our progression from spatial understanding to embodied control, we introduce a staged training pipeline, as shown in \cref{fig:teaser}. In \emph{Stage 1}, we develop a 3D-aware vision-language model, \textbf{\mnamevlm}, which extends a pretrained VLM by learning spatial reasoning from non-robotic 2D images annotated with 3D cues. This stage establishes a perceptual backbone that encodes geometric relations while preserving the rich semantic priors of large-scale pretraining. In \emph{Stage~2}, we introduce an \textit{action expert} that maps the grounded visual-language representations to motor actions. This stage demands well-tuned vision-language-action modeling choices and a carefully-crafted multi-embodiment data processing strategy to learn precise low-level robot controls. Together, these stages bridge the gap between internet-scale 2D perception and embodied 3D interaction, progressively transforming passive spatial understanding into actionable behavior.

Unlike previous works that address the challenge of 3D knowledge for robot control~\citep{spiridonov2025generalist,qu2025spatialvla}, \mname demonstrates improvement on a foundation model level, with an end-to-end policy across multiple different environments and robots.
It is capable of achieving state-of-the-art robot control on multiple robot embodiments without requiring target evaluation environment fine-tuning.
Furthermore, \mname demonstrates how significant amounts of robot demonstration data can be ``replaced” by non-robotic 3D-annotated image data. In summary, our work makes the following contributions:
\begin{itemize}
    \item\textbf{\mnamevlm}: a VLM with \emph{control-inspired 3D capabilities} (\eg localizing objects in 3D), trained on carefully-crafted VQA tasks and enriched 2D-image non-robotic data. Importantly, \mnamevlm directly boosts downstream VLA performance.
    \item\textbf{\mname}: an open-weight \emph{robotics foundation model with 3D understanding}, which significantly outperforms or matches the strongest state-of-the-art baselines, trained with 20$\times$ more robot demonstration data
    \item\textbf{Extensive experimental validation}: we demonstrate strong generalization across diverse settings with a substantial reduction in reliance on hard-to-collect robotic data. Notably, using only \textbf{200k non-robotic 2D images}, \mname surpasses models trained with more than \emph{900M additional frames of robotic demonstrations}.

\end{itemize}

\section{Related Work}
\label{sec:related_work}

\textbf{Spatial Understanding for VLMs}. The majority of existing VLMs trained on large-scale datasets have been limited to flat 2D image understanding~\citep{beyer2024paligemma, steiner2024paligemma, team2025gemma, karamcheti2024prismatic, wang2024qwen2, liu2023visualinstructiontuning}.
Our work extends the PaliGemma VLM \citep{beyer2024paligemma} by integrating the MoGe monocular depth estimator \citep{wang2024moge} as a supplementary vision backbone and by training on manipulation-relevant 3D tasks to enhance the VLM's 3D understanding.
Previously, \citet{chen2024spatialvlm} used a similar data annotation approach for training a 3D-aware VLM, but they do not integrate a pretrained depth estimator and neither their model nor their dataset is publicly accessible. Additionally, unlike SpatialVLM \citep{chen2024spatialvlm} or RoboSpatial~\citep{song2025robospatial}, trained on high-level spatial relationships, our \mnamevlm focuses on explicit 3D coordinate prediction, a pretraining task much closer to embodied control.
SpatialBot~\citep{cai2024spatialbot} has proposed a spatially-aware VLM targeting robot control, but their method involves a multi-step VLM inference process and was never shown to integrate into a VLA for generalist robotic control.

\noindent\textbf{Vision-Language-Action Models}. Recently, multiple works have developed generalist robot policies \citep{black2024pi_0, pertsch2025fast, intelligence2025pi_, kim2024openvla, dey2024revla, open_x_embodiment_rt_x_2023, zitkovich2023rt2} trained on multiple robot embodiments.
\mname builds on top of the $\pi_0$ architecture, which combines a pretrained PaliGemma VLM and an action expert module, but we initialize the underlying VLM from our \mnamevlm to integrate pretrained 3D understanding.
Previously, SpatialVLA~\citep{qu2025spatialvla} proposed integrating a monocular depth encoder \citep{depthanythingv2-2024} in the VLA, but without any VLM alignment or pretraining and therefore learning 3D capabilities entirely from hard-to-collect robotic data.
MolmoAct \citep{lee2025molmoact} recently proposed a spatially-aware VLA, but the approach involves 'reasoning' at inference time, rendering the method too slow for real-time control.
Most closely related, Gemini Robotics 1.0 \citep{team2025gemini} follows a similar 3D pretraining method to fine-tune the significantly larger Gemini 2.0 \citep{gemini2} and distill into a smaller VLA with reasoning capabilities. With most of the method's details undisclosed, our work still differs in several important aspects: (1) we investigate the benefits of 3D pretraining in isolation, (2) train much smaller open-access model on limited, less diverse open data from OpenX \cite{open_x_embodiment_rt_x_2023}, and, most importantly, (3) we demonstrate the ability to reduce the need for robotic data with non-robotic 2D images.

\section{Method}
\label{sec:method}

In this section, we describe \mname and its training recipe in detail.
In section \ref{subsec:3d-vlm} we describe the architecture, data generation pipeline, and training procedure of our 3D-aware \mnamevlm.
This stage aims to enhance the 3D spatial understanding capabilities of an off-the-shelf VLM through fine-tuning on 3D spatial perception tasks.
We then proceed, in section \ref{subsec:vla} to detail the architecture and training procedure of \mname, which comprises a pre-training and post-training stage.

\begin{figure*}
    \centering
    \includegraphics[width=\linewidth]{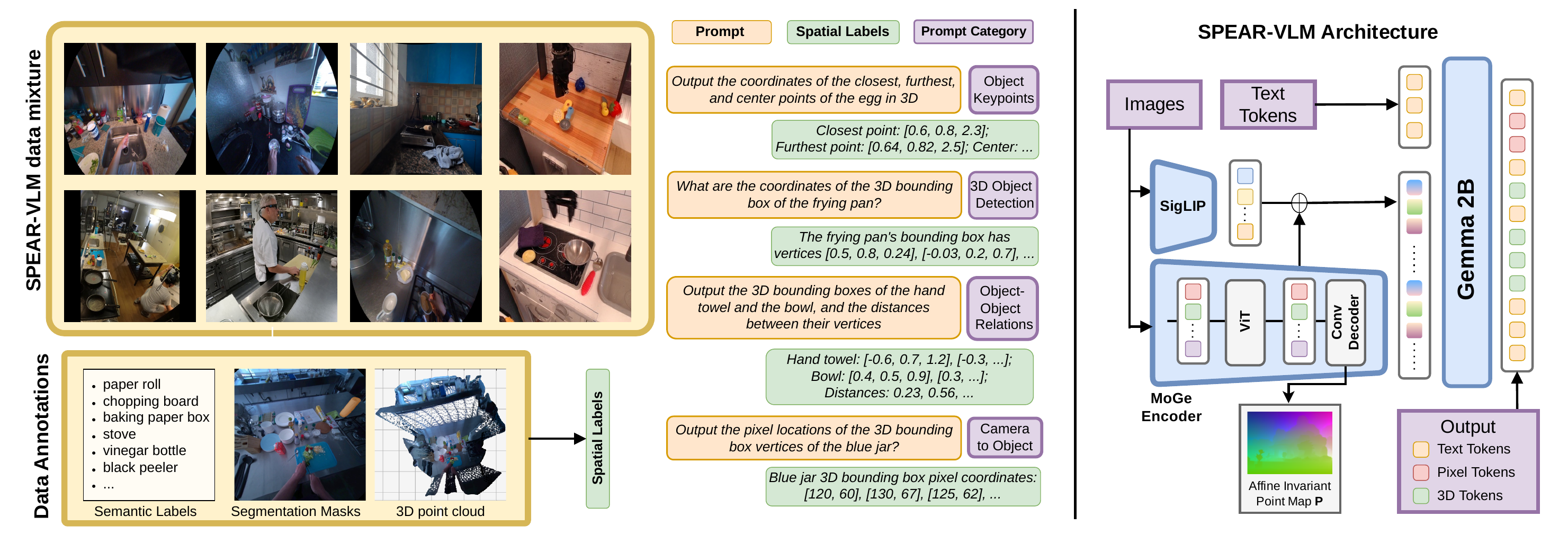}
    \caption{
        \textbf{\mnamevlm overview.}
        Left: Training data mixture, auto computed spatial annotations and example question-answer pairs from each category.
        Right: High-level architecture with fusion between SigLIP and MoGe encoders and PaliGemma embeddings expansion with 3D tokens. This design equips SPEAR-VLM with explicit 3D understanding that serves as a strong foundation for SPEAR-VLA.
    }
    \label{fig:3dvlm}
    \vspace{-7pt}
\end{figure*}
\subsection{\mnamevlm}
\label{subsec:3d-vlm}

Our approach considers the architecture of recent robotics foundational models that are based on VLMs, pretrained on large corpora of internet-scale text-image data
The architecture of those models usually consists of a vision encoder, a vision-to-text-embedding feature projector, and a LLM. The majority of the tasks on which VLMs are usually trained are limited to 2D \citep{beyer2024paligemma,  llava, karamcheti2024prismatic}.
To extend the capabilities of a pretrained VLM to 3D understanding, we propose (1) extending the model architecture by adding a monocular depth encoder and (2) training the VLM on VQA tasks that require explicit 3D reasoning.

\noindent\textbf{SPEAR-VLM Architecture}. Our model builds on PaliGemma \citep{beyer2024paligemma} as backbone, but the same method can be used with any late-fusion VLM \cite{Alayrac2022FlamingoAV, liu2024visual, driess2023palme}.
PaliGemma consists of three main components: (1) a SigLIP \textbf{visual encoder} \citep{zhai2023siglip}, (2) a linear \textbf{projector} that maps the visual tokens predicted by the visual encoder to the language model input space and (3) a Gemma \textbf{language model} \citep{team2024gemma}.
To enable the model to perceive accurate depth, we integrate the MoGe \citep{wang2024moge} depth encoder as an additional vision encoder.
We choose MoGe due to its affine-invariant modeling approach, capable of outputing 3D point clouds or depth, even \emph{when camera intrinsics are unknown}
Similar to the MoGe decoder inputs, we concatenate the intermediate features from the last 4 layers of the MoGe ViT encoder along the feature dimension and project them to the LLM embedding space via a randomly-initialized linear projector. The visual input to the LLM consists of the averaged outputs of the SigLIP and MoGe projectors. To encode 3D information into text we extend the PaliGemma tokenizer with $N = 1024$ 3D tokens (see \cref{fig:3dvlm} and Appendix \ref{subsec:app-depth-tokens}).

\noindent\textbf{3D pretraining tasks}. Given the above architecture, we propose a pre-training scheme to enable the model to leverage the depth information in MoGe's encoder features and acquire 3D spatial understanding capabilities. To embed as much control-relevant 3D knowledge in the \mnamevlm as possible, we design VQA tasks inspired by the embodied tasks a VLA needs to learn, \eg \textit{'Output the vertices of the 3D bounding box of object X'} or \textit{'Output the $xyz$ components of the distance between object X and object Y'}. These VLM pre-training tasks ensure learning semantic 3D localization, object-to-object spatial relations, and 3D coordinate system geometry (\cref{fig:3dvlm}). For a full list of question-answer pairs, see Appendix \ref{subsec:app-vlm-vqa-tasks}.

\noindent\textbf{3D Vision-Question-Answering Data}. As few open datasets contain the annotations needed for the proposed training scheme, we devise a semi-automatic annotation pipeline to enrich existing datasets with the necessary annotations: \textit{object-level segmentation masks, semantic labels and projected 3D point cloud}. Notably, our pipeline requires only 2D images as input and off-the-shelf vision foundation models:
\begin{enumerate}
    \item Use Gemini \cite{gemini2.5} to detect 2D bounding boxes and semantic labels for the objects in the image.
    \item Prompt SAM2 \citep{ravi2024sam, ren2024grounded} with the detected bounding boxes to produce instance-level segmentation masks.
    \item Obtain 3D point cloud annotations for the entire image via MoGe direct point cloud predictions \citep{wang2024moge}.
\end{enumerate}

To construct a training example, we randomly sample a templated text prompt and a set of objects from the image.
We then filter the annotated MoGe 3D point cloud with the object mask to obtain the object 3D point cloud.
Based on the segmented point cloud, we compute the oriented 3D bounding box and construct the question-answer pair.

We focus on indoor environments and annotate the ``cooking" and "bike repair" parts of EgoExo4D \citep{grauman2024ego} that already have segmentation masks, resulting in 200k images.
For visual diversity, we further annotate 30k frames of the Bridge-V2 \citep{walke2023bridgedata} robot demonstration dataset, downsampled to 10\% in the VLM training data mixture.

\noindent\textbf{Training process}. Similar to LLaVa \citep{llava}, we train \mnamevlm in two stages. In the first stage, we initialize from PaliGemma and MoGe weights, with the MoGe projector and the LLM 3D token embeddings initialized randomly. We train only the randomly initialized weights and SigLIP projector, keeping everything else frozen. In the second and longer stage, we keep only SigLIP and MoGe encoders frozen and we scale the next-token-prediction loss for 3D tokens by a factor $\lambda = 2$.

\subsection{\mname}
\label{subsec:vla}
\mname follows a similar overall architecture as $\pi_0$~\citep{black2024pi_0}, however, we build on \mnamevlm , use a rotation formulation in flow matching on the $\mathbb{S}^3$ manifold of unit quaternions, and several data \& engineering improvements. Design decisions were ablated on small-scale experiments on BridgeData V2~\citep{walke2023bridgedata} due to the cost of training on the entire OpenX mixture. We summarize these key decisions and learning in the following.

\noindent\textbf{Preliminaries.} Formally, we aim to learn a function $\pi(\cdot)$ mapping an observation $\mathbf{o}_t$  to a sequence of robot actions $\mathbf{A_t} = [\mathbf{a}_t, \mathbf{a}_{t+1}, \dots \mathbf{a}_{t+H-1}]$ over a horizon  $H$.
The observation is defined as $\mathbf{o}_t = [\mathbf{I}_{t}^{1}, \dots, \mathbf{I}_{t}^{n} , \mathbf{p}_t, \mathbf{l}_t]$, where $\mathbf{I}_t^{i}$ is the $i$-th image observation from an uncalibrated camera, $\mathbf{p}_t$ is a vector containing the robot state comprising of the end-effector pose and gripper state, $\mathbf{l}_t$ is a vector of language tokens representing the language instruction.

\noindent\textbf{Architecture.} We follow the broadly accepted architecture introduced in $\pi_0$: a Flow Matching action expert that processes proprioception observations and predicts the robot actions by attending to the VLM's intermediate key-value pairs.
For full details, see Appendix \ref{subsec:app-vla-arch} and $\pi_0$~\citep{black2024pi_0}.

\noindent\textbf{Flow Matching Formulation.} The action sequence prediction is supervised via conditional flow matching \citep{lipmanflow, liu2022rectified, lipman2024flow}.
The model takes as input the observation $\mathbf{o}_t$, the flow-matching step $\tau \in [0, 1]$ and a sequence of noisy actions $\mhat{\mathbf{A}}^\tau_t = [\mhat{\mathbf{a}}^{\tau}_{t}, \dots, \mhat{\mathbf{a}}^{\tau}_{t+H-1}]$  and predicts a denoising vector $\mathbf{v}_\theta(\mhat{\mathbf{A}}^{\tau}_t, \mathbf{o}_t)$.
We denote the decomposed action of translation, rotation and gripper components as $\mathbf{a}_t = [\mathbf{x}_t, \mathbf{q}_t, \mathbf{g}_t]$. We use the square brackets operator $[\cdot]$ on the predicted denoising vector $\mathbf{v}_\theta$ and the denoising vector field $\mathbf{u}$ to denote a specific component, \eg $\mathbf{u}[\mathbf{x_t}]$ corresponds to the translation component of the denoising vector field.

We follow a flow matching formulation in linear space for translation and on the $\mathbb{S}^3$ manifold of unit quaternions for rotation.
For simplicity, we omit the gripper component as it follows the same linear formulation as translation.

During training, we sample a random timestep $\tau \sim \mathcal{B}(\alpha, \beta) $ and random noise
$\mathbf{x}_\epsilon \sim \mathcal{N}(\mathbf{0}, \mathbf{I}), \mathbf{q}_\epsilon \sim \mathcal{U}(\mathbb{S}^3)$.
``Noisy actions" are computed by linear interpolation for translation $\mhat{\mathbf{x}}^\tau_t = \tau \mathbf{x}_t + (1 - \tau) \mathbf{x}_\epsilon$ and spherical linear interpolation on the
$\mathbb{S}^3$ manifold for quaternion rotation
\begin{equation}
\label{eq:slerp}
\mhat{\mathbf{q}}_t^\tau =
\frac{\sin\!\big((1 - \tau)\theta\big)}{\sin \theta}\,\mathbf{q}_\epsilon \;+\;
\frac{\sin(\tau \theta)}{\sin \theta}\,\mathbf{q}_t,
\end{equation}
with $\theta = \cos^{-1}(\mathbf{q}_\epsilon \cdot \mathbf{q}_t$).
The ``noisy action sequence" $\mhat{\mathbf{A}}^\tau_t$ is passed as input to the model and trained to output the denoising vector field
$\mathbf{u} (\mhat{\mathbf{A}}^\tau_t | \mathbf{A}_t) = \dfrac{d \mhat{\mathbf{A}}^\tau_t}{d \tau}$.
Training is supervised with the conditional flow-matching loss, equivalent to the MSE loss for translation
\begin{equation}
\mathcal{L}_{\mathbb{R}^3} (\theta) =
\big|\big|\mathbf{v}_\theta(\mhat{\mathbf{A}}^{\tau}_t, \mathbf{o}_t)[{\mathbf{X}_t}]
- \mathbf{u}(\mhat{\mathbf{A}}^{\tau}_t | \mathbf{A}_t)[{\mathbf{X}_t}]\big|\big|^2.
\end{equation}
For rotations, we combine a cosine loss between the velocity predictions $\mathbf{v}_\theta(\mhat{\mathbf{A}}^{\tau}_t, \mathbf{o}_t)[\mathbf{q}] \in \mathbb{R}^4$ and the denoising vector field $\mathbf{u}(\mhat{\mathbf{A}}^{\tau}_t | \mathbf{A}_t)[\mathbf{q}]$.
and a geodesic loss \citep{geist2024rotation, hartley2013rotation} between a target quaternion $\mhat{\mathbf{q}}_{t}^{\tau + \delta} \in \mathbb{S}^3$ computed from \cref{eq:slerp} at time $t + \delta$, and a quaternion prediction
$\mhat{\mathbf{q}}_{\theta, t}^{\tau + \delta} = \mhat{\mathbf{q}}_t^\tau \otimes \mhat{\mathbf{q}}_{\theta,t}^{\delta} \in \mathbb{S}^3$,
with $\mhat{\mathbf{q}}_{\theta,t}^{\delta} \in \mathbb{S}^4$ computed by integrating
$\mathbf{v}_\theta[\mathbf{q}_t] \in \mathbb{R}^4$
over a small integration step $\delta \sim \mathcal{U}(0.01, 1- \tau)$.
The total loss is the sum of the translation and rotation loss
\begin{equation}
\mathcal{L} (\theta) = \mathbb{E}_{p(\mathbf{A}_t| \mathbf{o}_t), q(\mhat{\mathbf{A}}^{\tau}_t | \mathbf{A}_t)}
\left[
\mathcal{L}_{\mathbb{R}^3} (\theta) + \mathcal{L}_{\mathbb{S}^3} (\theta)
\right].
\end{equation}

During inference, we generate actions by integrating the learned vector field from $\tau = 0$ to $\tau = 1$, starting with random noise
$\mhat{\mathbf{x}}_0 \sim \mathcal{N}(\mathbf{0}, \mathbf{I}), \mhat{\mathbf{q}}_0 \sim \mathcal{U}(\mathbb{S}^3)$
and using Euler integration in linear space for translations
\begin{equation}
    \mhat{\mathbf{x}}^{\tau + \delta}_t = \mhat{\mathbf{x}}^{\tau}_t + \delta \mathbf{v}_\theta^\mathbf{x} (\mhat{\mathbf{A}}^{\tau}_t, \mathbf{o}_t),
\end{equation}
and on the $\mathbb{S}^3$ manifold for rotations
\begin{equation}
    \mhat{\mathbf{q}}^{\tau + \delta}_t = \mhat{\mathbf{q}}^{\tau}_t \otimes \mhat{\mathbf{q}}_{t}^{\delta} (\mathbf{v}_\theta^\mathbf{q} (\mhat{\mathbf{A}}^{\tau}_t, \mathbf{o}_t)).
\end{equation}
See Appendix \ref{subsec:app-vla-fm} for more details on flow matching.

\noindent\textbf{Image Resolution.}
We select a resolution of $280\times210$ for the main external camera and $112\times112$ for the wrist camera, and resize images either by a central crop or padding.
Importantly, unlike prior work~\citep{kim2024openvla}, we do not distort the aspect ratio of the images by naive resizing as this also distorts camera intrinsics and negatively affects depth and point cloud estimates.
As wrist cameras contain less information than the external camera, we use a lower resolution, without losing important details, and reduce training and inference compute.

\noindent\textbf{Fine-tuning vision encoders.}
As previously observed by ReVLA~\citep{dey2024revla}, robotics training can degrade the representations of the pre-trained vision encoders.
We experiment with various configurations of vision encoder training and find the optimal setting to keep both SigLIP and MoGe vision encoders trainable during VLM training, but freeze MoGe in the VLA training stage.

\noindent\textbf{Control frequency \& Data normalization.}
We use an action chunk of size $H=5$ and frequency of 5Hz.
For datasets not providing observations at 5Hz we resample the action targets via linear interpolation.
We design data normalization to encourage learning motion across datasets, instead of “memorizing” each dataset separately.
For target control normalization, we use global quantile normalization with statistics computed across the entire training mixture.

\noindent\textbf{Rotations.}
We investigate various rotation representations including Euler angles, rotation matrices and unit quaternions.
This is run in combination of using different rotation losses, including MSE or cosine for velocity predictions and geodesic and/or chordal loss \citep{hartley2013rotation} for integrated rotation predictions, as well as end-effector or robot base reference frames.
We use Gram-Schmidt orthonormalization~\citep{geist2024rotation} to ensure valid rotation matrix predictions, but we find half-space unit quaternions to produce better results overall.
We also find our proposed formulation on the manifold of unit quaternions $\mathbb{S}^3 \rightarrow \mathbb{S}^3$ to be more stable and effective than linear flow matching $\mathbb{R}^4 \rightarrow \mathbb{S}^3$.

\noindent\textbf{Evaluation and Checkpointing.}
We ablate all design choices by evaluating on the SIMPLER WidowX environments~\citep{li24simpler}.
We set the same seed and enable deterministic CUDA operations for all VLA ablations to reduce training variance. We further resort to exponential moving average (EMA) checkpointing, which significantly stabilizes final checkpoint performance. For further details and ablations, see Appendix \ref{subsec:app-vla-design-ablate}.
\section{Experimental evaluation}
\label{sec:exp-eval}

\begin{figure*}
    \centering
    \includegraphics[width=0.9\linewidth]{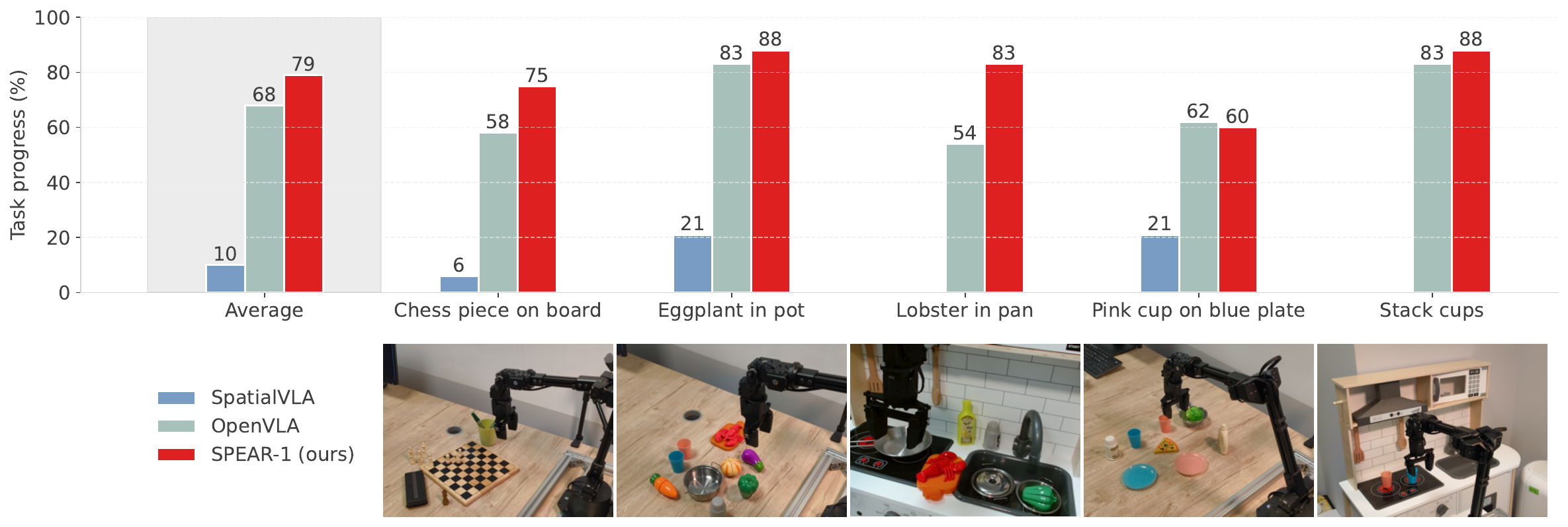}
    \caption{
        \textbf{Real world evaluation on WidowX.}
        \mname is able to achieve 10\% higher average task progress across all tasks than OpenVLA, a strong baseline in this setting. Bottom images correspond to the real-world tasks, whose performances are reported above.
    }
    \label{fig:realworld-widowx}
    \vspace{-10pt}
\end{figure*}

\begin{figure*}[!t]
    \centering
    \includegraphics[width=0.9\linewidth]{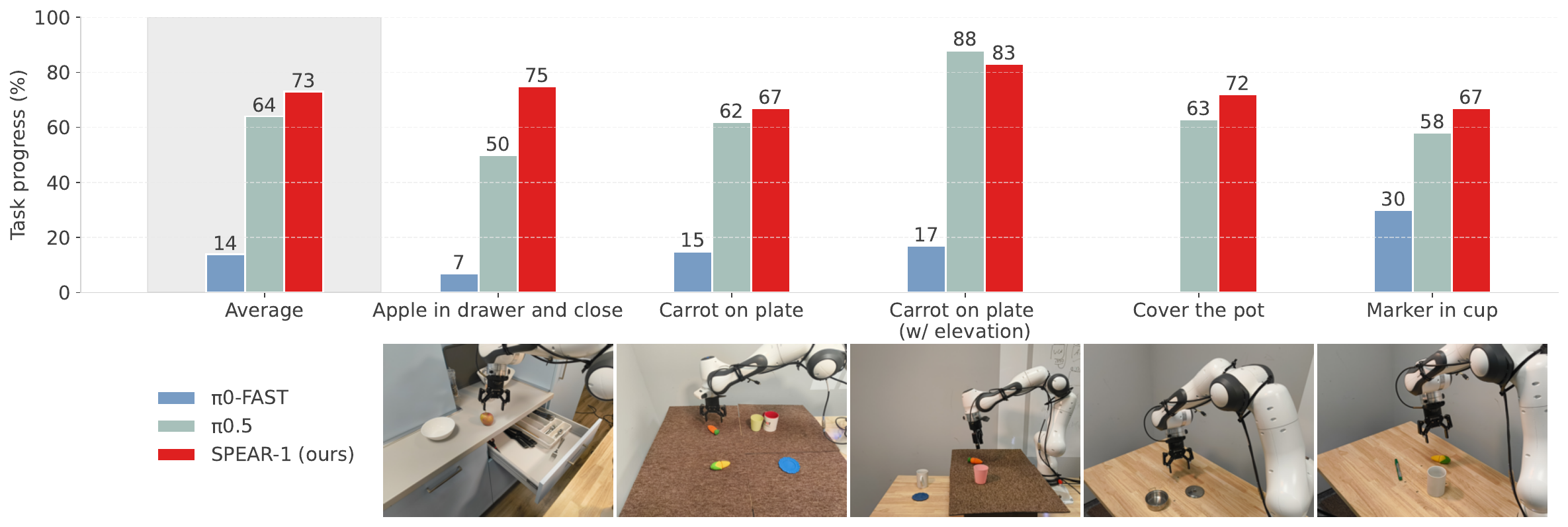}
    \caption{
        \textbf{Real world evaluation on Franka.}
        We find that without any fine-tuning on the target environment, \mname noticeably outperforms $\pi_0$-FAST, and matches $\pi_{0.5}$, even though both baselines are trained on 20$\times$ more robotic data from significantly more diverse environments. The bottom row shows challenging, varied Franka environments where SPEAR-1 maintains strong zero-shot performance.
    }
    \vspace{2mm}
    \label{fig:realworld-franka}
    \vspace{-12pt}
\end{figure*}

We evaluate the performance of \mname as a generalist policy for robot manipulation and compare it to open-weights state-of-the-art VLA models. Our experiments aim to answer the following research questions:

\begin{enumerate}
    \item Does 3D VLM pretraining improve the downstream VLA performance on robot control tasks?
    \item How well does \mname compare against state-of-the-art VLA models?
\end{enumerate}

To answer these questions, we evaluate \mname on a variety of manipulation tasks in simulation and multiple real-world environments on several robot embodiments.

\subsection{Implementation details}
\label{subsubsec:impl-details}

\textbf{VLM training.} We train \mnamevlm with a batch size of 512 for 2k steps during the first stage and 10k steps for the second, for a total of ~18hrs on 16 Nvidia H200 GPUs.\\
\textbf{VLA pre-training.} For VLA training, we start from \mnamevlm and randomly initialized action expert.
We provide two camera views as inputs to the model: external, with resolution 280x210, and wrist, with resolution 112x112.
When the wrist camera is not available, we feed a black image.
We train on 32 H200 GPUs with batch size 2048 for 300k steps ($\sim$6 days) on a data mixture comprising 24 datasets (see Appendix \ref{subsec:app-vla-data}) from the Open X-Embodiment (OXE) collection~\cite{open_x_embodiment_rt_x_2023}.\\
\textbf{VLA post-training.} For WidowX real-world and SIMPLER simulation and Franka real-world experiments, we additionally fine-tune our OXE pre-trained \mname for 50k steps on the Bridge V2~\citep{walke2023bridgedata} and DROID~\citep{khazatsky2024droid} datasets respectively.
We refer to these versions as \textbf{\mname (Bridge)} and \textbf{\mname (DROID)} respectively.

\begin{table*}[!ht]
  \centering
  \resizebox{\textwidth}{!}{
  \begin{tabular}{l|l|c|cc|cc|cccc|c}
    \toprule
    Experiment   & \shortstack{VLM\\Architecture}  & 3D tasks     & \multicolumn{2}{c|}{\shortstack[c]{VLM training\\[-0.15ex]\begin{tabular}{@{}cc@{}} SigLIP & MoGe \end{tabular}}} & \multicolumn{2}{c|}{\shortstack[c]{VLA training\\[-0.15ex]\begin{tabular}{@{}cc@{}} SigLIP & MoGe \end{tabular}}}  & \shortstack{Put Carrot\\on Plate} & \shortstack{Put Eggplant in\\Yellow Basket} & \shortstack{Put Spoon\\on Towel} & \shortstack{Stack Green Block\\on Yellow Block} & \shortstack{Avg. Success\\Rate} \\
    \midrule
    1. no 3D     & \textbf{PaliGemma}  & None              & train           & --              & train   & --              & 25\%    & 0\%    & 54\%    & 4.1\%  & 20.8\% \\
    2. no OBJ    & \textbf{\mnamevlm}  & \textbf{points}   & frozen          & frozen          & train   & frozen          & 37.5\%  & 0\%    & 45.8\%  & 4.1\%  & 20.8\% \\
    3. no MoGe   & \textbf{PaliGemma}  & \textbf{objects}  & frozen          & --              & train   & frozen          & 29.1\%  & 0\%    & 70.8\%  & 4.1\%  & 26.0\% \\
    3. no VLM-T  & \mnamevlm           & \textbf{objects}  & frozen          & frozen          & train   & frozen          & 41.7\%  & 0\%    & 70.8\%  & 0\%    & 29.1\% \\
    4. no VLA-MF & \mnamevlm           & objects           & frozen          & frozen          & train   & \textbf{train}  & 29.1\%  & 0\%    & 41.7\%  & 4.1\%  & 18.8\% \\
    5. \mnamevlm & \mnamevlm           & objects           & \textbf{train}  & \textbf{train}  & train   & frozen          & 50\%    & 0\%    & 79.1\%  & 12.5\% & \textbf{35.4\%} \\
    \bottomrule
  \end{tabular}
  }
\caption{
    \textbf{\mnamevlm 3D ablations}.
    Ablations on a single environment subset of Bridge V2~\citep{walke2023bridgedata} to show the impact of 3D VLM pretraining.
    Without object-level 3D tasks (OBJ), 3D VLM pretraining does not show improvement over PaliGemma (1 vs. 2).
    Object-level 3D tasks without MoGe shows some, but not significant improvement over PaliGemma (1 vs. 3).
    Training MoGe during VLA training (VLA-MF) significantly degrades performance (4 vs. 5).
    The optimal training configuration is obtained when both SigLIP and MoGe are trained during VLM pretraining (6), followed by the frozen MoGe during VLA training.
}
\vspace{-5mm}
\label{tab:vlm-3d-ablation}
\end{table*}

\begin{table}[!t]
\centering
\small
\vspace{4pt}
\resizebox{\columnwidth}{!}{
\begin{tabular}{l|ccc|c}
\toprule
Method & \shortstack{Carrot on\\Plate (Dist)} & \shortstack{Carrot on\\Plate (Elev.)} & \shortstack{Marker in\\Cup (Dist)} & \shortstack{Avg. Task \\Progress}\\
\midrule
$\pi_0$-PaliGemma (DROID) & 0\% & 32\% & \textbf{67\%} & 34\% \\
$\pi_0$-\mnamevlm (DROID) & \textbf{42\%} & \textbf{52\%} & 43\% & \textbf{46\%} \\
\bottomrule
\end{tabular}
}
\caption{
    \textbf{SPEAR-VLM vs. PaliGemma for the downstream VLA tasks.}
    The experiments were conducted on the Franka (DROID) platform and the models were from trained from scratch on DROID.
    \mnamevlm achieves noticeable improvements.
    "Carrot on Plate" is not a part of DROID. This indicates \mnamevlm leads to better generalization.
}
\vspace{-5mm}
\label{tab:vlm-ablation}
\end{table}

\subsection{3D ablations: \mnamevlm vs PaliGemma}
\label{subsec:ablation_paligemma}

We first evaluate whether 3D VLM pretraining improves VLA performance on downstream tasks and what design choices matter. Due to the cost of training on the entire OXE mixture (\cref{tab:vla-data-mixture}), we train only on specific subsets.

\noindent\textbf{SIMPLER WidowX experiments}.
We perform an ablation study by training on a subset of the Bridge V2~\citep{walke2023bridgedata} dataset, containing demonstrations only from a single kitchen sink environment, and evaluate the models in the SIMPLER~\citep{li24simpler} WidowX environments (see Appendix \ref{subsec:app-vla-3d-ablate} for details).
This induces a distribution shift, which allows to demonstrate the benefits of 3D pretraining when evaluating in unseen environments.
In contrast, training on entire Bridge V2 leads to nearly the same performance for all models due to the close match between training and evaluations.

Results are reported in \cref{tab:vlm-3d-ablation}.
First, we note that simply using \mnamevlm architecture and training without object-level prompts, but only 3D coordinates of random pixels (row 2), does not lead to any meaningful change in VLA performance over the baseline $\pi_0$ model based on PaliGemma (row 1).
Training PaliGemma without MoGe encoder on all 3D object-level tasks (\cref{fig:3dvlm}), leads to some improvement (row 3 vs 1).
Integrating MoGe leads to an even bigger jump in performance (row 6 vs. row 3).
We also observe the importance of training SigLIP and MoGe encoders both during VLM and VLA training (row 4-6), with the best performance achieved when both are fine-tuned during VLM training and frozen MoGe during VLA training (row 6).
We hypothesize this is because SigLIP has been trained only for image level semantics, while MoGe has been trained for dense and detailed depth prediction, which is much closer to the nature of robotic manipulation.

\noindent\textbf{Real-world Franka experiments}.
To further validate the benefits of 3D VLM pretraining, we run comparisons by training on the DROID dataset~\citep{khazatsky2024droid}.
Due to the higher cost, we train only 2 models: one initialized from the base PaliGemma VLM and the other from our 3D-aware \mnamevlm.
We refer to the resulting models as $\pi_0$-PaliGemma (DROID) and $\pi_0$-\mnamevlm (DROID) respectively.
We compare the performance of both VLAs on three of the four tasks from our Franka experiments (Section \ref{subsubsec:realworld-exp}).
The results are reported in \cref{tab:vlm-ablation}.
We can observe that $\pi_0$-\mnamevlm (DROID) is able to outperform the baseline by more than 10\% on average.
We note that the task ``Carrot on plate" is not a part of the DROID training dataset, thus shows the improved generalization capabilities of \mnamevlm.
The lower scores of both models on the variations tabletop/elevations are likely due to workspace 3D position being out-of-distribution. Even in this case, $\pi_0$-\mnamevlm (DROID) is able to successfully complete the task in some cases while $\pi_0$-PaliGemma (DROID) fails every time.

\subsection{Simulation experiments}
\label{subsubsec:sim-exp}

We evaluate \mname on the WidowX environments of the SIMPLER simulation benchmark \cite{li24simpler}, and compare it with OpenVLA \cite{kim2024openvla} and SpatialVLA \cite{qu2025spatialvla}.
We report the results in \cref{tab:simplerenv_widowx}. Our model is able to outperform the baselines by more than 10\%.
In our experience, we found SIMPLER simulation results only to be indicative of relative performance of the models on the real WidowX robot, but not necessarily of absolute performance. Therefore, we focus on real-world evaluations.

\begin{table}[!ht]
  \centering
  \resizebox{\columnwidth}{!}{
  \begin{tabular}{l|cccc|c}
    \toprule
    Model                     & \shortstack{Put Carrot\\on Plate} & \shortstack{Put Eggplant\\in Yellow Basket} & \shortstack{Put Spoon\\on Towel} &  \shortstack{Stack Green Block\\on Yellow Block} & \shortstack{Avg. Success\\Rate} \\
    \midrule
    OpenVLA                   & 0\%    & 4.1\%   & 0\%    &  0\%     & 1.0\% \\
    SpatialVLA                & 25.0\% & \textbf{100.0}\% & 16.7\% &  29.2\%  & 42.7\% \\
    \textbf{\mname (ours)}    & \textbf{58.3}\% & 62.5\%  & \textbf{62.5}\% &  \textbf{45.82}\% & \textbf{57.3\%} \\
    \bottomrule
  \end{tabular}
  }
  \caption{
    \textbf{SIMPLER~\citep{li24simpler} simulation evaluations.}
    SpatialVLA numbers are  from~\citep{qu2025spatialvla}.
    SPEAR-1 outperforms the considered baselines by more than 10\%.
  }
  \vspace{-5mm}
  \label{tab:simplerenv_widowx}
\end{table} 

\subsection{Real-world experiments}
\label{subsubsec:realworld-exp}

We conduct evaluations on a total of 10 manipulation tasks across two robot platforms: WidowX and Franka Research 3. The tasks are designed to assess the ability of the evaluated models to generalize to unseen environments and objects. We design the tasks to be challenging for all models. For more details about the selected tasks see Appendix \ref{subsec:app-robot-tasks-scoring-rubrics}.

\noindent\textbf{Evaluation protocol}. For each task we define M initial conditions by varying the starting position of the objects in the scene. We execute N trials for each initial condition, for a total of N$\times$M trials per task. For each model, we evaluate and report the average task progress across all tasks, configurations, and trials. To that end, following previous works \cite{open_x_embodiment_rt_x_2023, barreiros2025lbm}, we define a scoring rubric with partial scoring for each task (see Appendix \ref{subsec:app-robot-tasks-scoring-rubrics} for scoring rubrics details).

\noindent\textbf{WidowX experiments}. Our hardware setup for this set of experiments closely matches the Bridge V2 setup \citep{walke2023bridgedata}, with a single external camera positioned on the side of the robot arm, pointing toward the workspace. For this set of experiments, 5 tasks are evaluated, with M = 4, N = 3, for a total of 60 trials per model. We compare the performance of \mname with OpenVLA \citep{kim2024openvla}, using the publicly released implementation and model weights.
In this setting, we do not compare against $\pi_0$~\citep{black2024pi_0}, $\pi_0$-FAST~\citep{pertsch2025fast} and $\pi_{0.5}$~\citep{intelligence2025pi_} due to the lack of publicly accessible weights for the WidowX platform.
The results are reported in \cref{fig:realworld-widowx}. \mname is able to achieve 10\% higher average task progress across all tasks than OpenVLA, a very strong baseline in this setting.

\noindent\textbf{Franka experiments.} Our hardware setup for this set of experiments is similar to that of DROID \citep{khazatsky2024droid}.
We design 5 tasks, with M = 5 and N = 3, for a total of 75 trials per model.
We found that the wrist camera view is crucial for training and deployment on DROID.
To ensure a fair comparison, we compare against open-weights models that use both the external and wrist camera.
Specifically, we compare \mname with the DROID-finetuned variants of $\pi_0$-FAST \citep{black2024pi_0,pertsch2025fast}, a strong autoregressive baseline, and $\pi_{0.5}$ \citep{intelligence2025pi_}, one of the latest state-of-the-art robotic foundation model optimized for open-world generalization.

The results of our real-world experiments are reported in \cref{fig:realworld-franka}. Without any fine-tuning on the target environment, \mname is able to significantly outperform $\pi_0$-FAST, and match $\pi_{0.5}$.
We note that both baselines do not integrate any sort of specialized 3D-aware training and are trained on at least 900M more robot demonstration frames collected in diverse environments. In contrast, \mname is trained on $\sim$45M frames, approximately $20\times$ less robotics data.
These results indicate the importance of 3D-based knowledge and pretraining for VLA's generalization.
As an architecturally close comparison, $\pi_0$-FAST integrates a specialized action tokenization compared to $\pi_0$ and was the first generalist policy trained on the DROID~\citep{khazatsky2024droid} to be successfully evaluated zero-shot in unseen environments, without fine-tuning.
In comparison, \mname, which also follows the $\pi_0$ architecture, can reach $\sim$$5$$\times$ higher performance than $\pi_0$-FAST without fine-tuning and without the large-scale robotic data used by $\pi_0$-FAST.

Apart from architectural enhancements and co-training on top of $\pi_0$-FAST, $\pi_{0.5}$ integrates high-level semantic subtask prediction and robotic data mixture explicitly focused on environment diversity and generalization.
Qualitatively and quantitatively, we find $\pi_{0.5}$ to perform better at environment generalization than $\pi_0$-FAST and match \mname's performance on our set of evaluation tasks.
This suggests that 3D VLM pretraining on non-robotic data from diverse environments is a more scalable way to boost robotic models' generalization capabilities without the need for large-scale robotic data collection in diverse environments.

\section{Discussion and Limitations}
\label{sec:discussion}

As highlighted by our experimental evaluation, SPEAR-1 achieves state-of-the-art performance in multiple zero-shot robot control scenarios, both in simulation and in the real-world.
Nevertheless, our approach has some limitations.
The proposed 3D VLM pre-training strategy is not well suited for deformable objects or objects with complex shapes.

Future work could explore the use of different 3D priors such as rotation estimation and collision detection to better capture the geometry of such objects.
Additionally, the coordinates of the 3D bounding boxes labels computed using MoGe's affine-invariant depth predictions are not in metric space. Further investigation is required to analyze the implications of this design choice on downstream performance, as well as to explore how ground truth point cloud labels or state-of-the-art metric-depth estimators could be integrated to help resolve this limitation.

While we have showed the benefits of 3D VLM pre-training on downstream robot control tasks, the scaling laws relating the latter to the quantity and quality of 3D pre-training data are still not well understood. Due to resource and time constraints, we leave this investigation for future work.
Another limitation of SPEAR-1 is the need to fine-tune on the target embodiment to achieve satisfactory results. We plan to explore how to alleviate this requirement in future work.
It also remains to be seen how well \mname generalizes to orders of magnitude more tasks and environments against models such as $\pi_{0.5}$ trained on significantly more diverse robot data.

\section{Conclusion}
\label{sec:conclusion}
In this work we introduced \mname and \mnamevlm that demonstrate a path towards building generalist robot policies from data beyond robot teleoperation only.

Our method targets the VLM backbone with \textbf{\mnamevlm}, a 3D-aware VLM trained on 2D images from non-robotic datasets enriched with 3D annotations. To embed control-relevant 3D knowledge in \mnamevlm, we train it on VQA questions, inspired by embodied tasks.
Stepping on this foundation, we built \textbf{\mname}, a robotic foundation model that can be deployed robustly across multiple robot platforms and environments, and matches or outperforms state-of-the-art foundation models which have been trained on $20\times$ more robot demonstrations.

Our work supports the hypothesis that enhancing VLM capabilities with non-robotic embodied knowledge is a scalable way to \textit{reduce dependence on hard-to-collect robot demonstrations and build future robotic foundation models}.

\section*{Acknowledgments}

{Project Lead:} Nikolay Nikolov, {Project Manager:} Jan-Nico Zaech, {PI:} Danda Pani Paudel, Luc Van Gool

We thank Alexander-Marc Spiridonov, Anna-Maria Halacheva, Yutong Hu for feedback and helpful technical discussions. We also thank Hristo Venev for engineering support and Kamen Pavlov for help with figures and visuals.

This research was partially funded by the Ministry of Education and Science of Bulgaria (support for INSAIT, part of the Bulgarian National Roadmap for Research Infrastructure).

{
    \small
    \bibliographystyle{ieeenat_fullname}
    \bibliography{main}
}

\clearpage
\appendix
\section{Appendix}
\label{sec:appendix}

The appendix is organized as follows:
\begin{itemize}
    \item In \cref{subsec:app-vlm-train} we provide more details on the VLM pre-training including VQA tasks, encoder fusion strategies, 3D tokenization and data annotation pipeline.
    \item In \cref{subsec:app-vla-train} we provide more details on the VLA training including data mixture, architecture, flow matching and design decision ablation results.
    \item In \cref{subsec:app-robot-tasks-scoring-rubrics} we provide the scoring rubrics for real-world evaluation tasks
    \item In \cref{subsec:app-zero-shot-perf} we discuss the differences between Bridge V2 and DROID datasets for zero-shot control evaluations in unseen environments.
\end{itemize}

\begin{table}[ht]
\centering
\small
\begin{tabular}{l r}
\toprule
\textbf{Dataset} & \textbf{Weight} \\
\midrule
austin\_buds\_dataset        & 0.5  \\
austin\_sailor\_dataset      & 2.0  \\
austin\_sirius\_dataset      & 0.5  \\
berkeley\_autolab\_ur5       & 1.0  \\
berkeley\_cable\_routing     & 0.1  \\
berkeley\_fanuc\_manipulation& 1.0  \\
bridge                       & 18.0 \\
dlr\_edan\_shared\_control   & 0.1  \\
droid                        & 35.0 \\
fmb                          & 1.5  \\
fractal20220817\_data        & 12.0 \\
furniture\_bench\_dataset    & 1.5  \\
iamlab\_cmu\_pickup\_insert  & 0.3  \\
kuka                         & 4.0  \\
language\_table              & 1.5  \\
nyu\_franka\_play\_dataset   & 0.3  \\
roboset (kinesthetic)        & 2.0  \\
roboset (teleoperation)      & 5.0  \\
roboturk                     & 3.0  \\
stanford\_hydra\_dataset     & 3.0  \\
taco\_play                   & 2.0  \\
toto                         & 1.5  \\
ucsd\_kitchen\_dataset       & 0.2  \\
utaustin\_mutex              & 3.0  \\
viola                        & 1.0  \\
\bottomrule
\end{tabular}
\caption{Open X-Embodiment data mixture for \mname pre-training}
\label{tab:vla-data-mixture}
\end{table}
\begin{table*}[!t]
\centering
\begin{tabular}{l l r c}
\hline
\textbf{Dataset} & \textbf{Domain / Subset} & \textbf{\# Annotated Images} & \textbf{Segmentation Masks} \\
\hline
EgoExo4D \citep{grauman2024ego} & Cooking \& Bike Repair & $\sim$200k & GT \\
Bridge \citep{walke2023bridgedata} & Robot Demonstrations   & $\sim$30k  & SAM2 Generated  \\
\hline
\textbf{Total} & & $\sim$230k & \\
\hline
\end{tabular}
\caption{Annotated image counts for training dataset construction, with segmentation mask availability.}
\label{tab:vlm-data-mixture}
\end{table*} %
\begin{table*}[t!]
\centering
\resizebox{\textwidth}{!}{
\begin{tabular}{p{4cm}cccc}
\toprule
\textbf{Task} & \textbf{0.25} & \textbf{0.50} & \textbf{0.75} & \textbf{1.00} \\
\midrule
Carrot on Plate\\ (w/ distractors \& elevations)
 & Reach carrot
 & Pick up carrot
 & Drop on/near plate
 & Correctly place on plate \\
\addlinespace
Marker in Cup\\ (w/ distractors)
 & Reach marker
 & Pick up marker
 & Drop on/near cup
 & Place inside cup \\
\addlinespace
Cover the Pot
 & --
 & Pick up lid
 & Drop lid on pot
 & Correctly cover pot \\
\addlinespace
Apple in drawer and close
& Pick up the apple 
& Put the apple in the drawer 
& Half-close the drawer
& Fully close the drawer \\
\bottomrule
\end{tabular}
}
\caption{Scoring rubric for Franka evaluation tasks.}
\label{tab:real_world_scoring}
\end{table*}

\begin{table*}[t!]
\centering
\resizebox{\textwidth}{!}{
\begin{tabular}{p{4cm}cccc}
\toprule
\textbf{Task} & \textbf{0.25} & \textbf{0.50} & \textbf{0.75} & \textbf{1.00} \\
\midrule
Eggplant in pot
 & Reach the eggplant
 & Pick up the eggplant
 & Drop the eggplant near the pot
 & Drop the eggplant on the pot \\
\addlinespace
Pink cup on blue plate
 & Reach the pink cup
 & Pick up the pink cup
 & Drop the pink cup near the plate
 & Place the pink cup correctly \\
\addlinespace
Chess piece on board 
 & - 
 & Go to the brown chess piece
 & Pick up the brown chess piece
 & Drop it on the board \\
\addlinespace
Lobster in the pan 
 & - 
 & Pick up the lobster
 & Drop the lobster near the pan
 & Place the lobster inside the pan \\
\addlinespace
Corn between cups
 & - 
 & Pick up the corn
 & - 
 & Place the corn between the cups \\
\bottomrule
\end{tabular}
}
\caption{Scoring rubric for WidowX evaluation tasks.}
\label{tab:real_world_scoring_widowx}
\end{table*}

\begin{figure*}
    \centering
    \includegraphics[width=\linewidth]{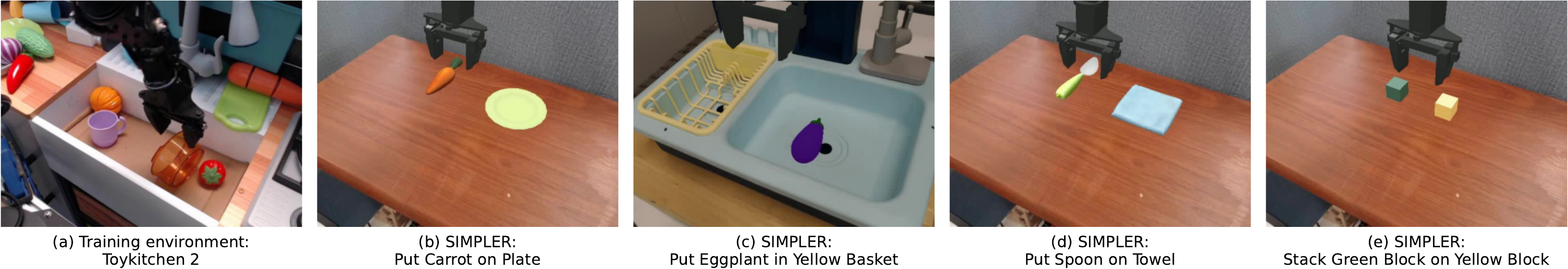}
    \caption{
        \textbf{3D ablation environments on WidowX.}
        (a) Training data subset from Bridge V2~\citep{walke2023bridgedata}.
        (b) - (e) SIMPLER evaluation environments.
    }
    \label{fig:3d-widowx}
\end{figure*}

\subsection{VLM training}
\label{subsec:app-vlm-train}

\subsubsection{VQA tasks for VLM pre-training}
\label{subsec:app-vlm-vqa-tasks}

The Visual Question Answering (VQA) tasks used during VLM pre-training are inspired by VLA embodied tasks and aim to embed as much control-relevant 3D information into the VLM as possible. We use templated question-answer pairs grouped in the following categories:

\begin{itemize}
    \item \textbf{3D keypoints prediction}: Output the 3D coordinates of the closest, furthest and center points of an object with respect to the camera frame.
    \item \textbf{3D bounding prediction}: Output the vertices of the 3D bounding box of an object.
    \item \textbf{Object-to-object distance prediction}: Output the direct distance between object X and object Y in 3D space as well as its $xyz$ components.
    \item \textbf{Object-to-object bounding box prediction}: Output the distance between the bounding box vertices and the centers of object X and object Y.
    \item \textbf{Backprojection}: Locate the vertices of the 3D bounding box of an object on the 2D image.
    \item \textbf{Chain-of-thought comparisons}: What is the distance from the camera to object X? What is the distance from the camera to object Y? Which object is closer to the camera?
\end{itemize}

To further encourage the model to 'reason' over the information provided and attend to the right objects, in a single training example we use a random number (between 1 and 4) of question-answer pairs corresponding to different prompts and objects in the scene. To resolve ambiguities, if two instances of the same type of object appear in the image, we filter them out and never ask questions about them.

\subsubsection{VLM encoder fusion strategies}
\label{subsec:app-encoder-feat-fusion}
We experimented with 2 different strategies to combine the outputs of the SigLIP and MoGe encoders:
\begin{enumerate}
    \item Averaging the visual features predicted by both encoders after projecting each set of features through a separate linear layer to the LLM embedding space. In particular, for SigLIP we take only the tokens at the last layer of the vision encoder, while for MoGe we take the tokens at the last 4 layers of the encoder, following the approach used by MoGe architecture to decode the features to a 3D point cloud.
    \item Using MoGe's predicted 3D point cloud $\mathbf{P}$ in the camera ego pose (in an affine-invariant space) and adding them to the SigLIP encoder features, similar to SpatialVLA \citep{qu2025spatialvla}. In particular, MoGe's 3D point cloud output $\mathbf{P} \in \mathbb{R}^{H \times W \times 3}$ is embedded to $\mathbf{P^\prime} \in \mathbb{R}^{ h \times w \times d}$ through a projector $\psi(\cdot)$, composed of normalization, convolution, sinusoidal embedding $\gamma(x) = (x, \sin(2^0\pi x), \cos(2^0 \pi x), \dots , \sin(2^{L-1} \pi x), \cos(2^{L-1} \pi x) )$ \citep{mildenhall2021nerf} and an MLP. Finally, the features $\mathbf{F^\prime} = \mathbf{F} + \mathbf{P^\prime}$ are fed to PaliGemma's SigLIP linear projector, where $\mathbf{F} \in \mathbb{R}^{h \times w \times d}$ denotes the features at the SigLIP encoder output.
\end{enumerate}

During our preliminary VLM evaluations we found the first strategy to demonstrate qualitatively better performance on bounding box prediction tasks. In particular, models trained with the second approach struggled to consistently output "grammatically" correct bounding boxes, \eg they would output 22 or 23 3D tokens instead of the required 24. We therefore used the first approach for all VLM pre-training experiments in the main paper.

\subsubsection{3D tokenization}
\label{subsec:app-depth-tokens}

To encode 3D information into text we extend the PaliGemma tokenizer with $N = 1024$ 3D tokens, as 3D coordinates are conceptually different from the existing visual and language tokens. This is in line with PaliGemma's approach of extending Gemma's tokenizer to pixel locations. Each 3D token corresponds to a quantized \textit{distance value} in the range $[z_{\min}, z_{\max}]$, where $z_{\min}$ and $z_{\max}$ are computed as the 1st and 99th quantiles of the 3D point cloud distribution along any of the $xyz$ coordinates.

We found the \textit{distance values} in the data to approximately follow a Normal distribution. Therefore, to allow for more accurate tokenization, we compute non-uniform bins with fine-grained discretization around the mean and spread out widths near the tails such that the distribution of 3D tokens is approximately uniform.

We initialize the new token embedding weights from a multivariate normal distribution that has the mean and covariance of the pretrained embeddings~\citep{mean_resizing, mean_resizing_hf}.

\subsubsection{VQA data annotation pipeline}
\label{subsec:app-vlm-vqa-data}

We follow the method described in Section~\ref{subsec:3d-vlm} in order to enrich 2D images with semantics, segmentation masks and 3D point clouds.
We also experimented with GroundingDINO~\citep{liu2023grounding} instead of Gemini, but we found the semantic labels produced by GroudingDINO to be a lot less accurate and consistent.
We found that prompting SAM2~\citep{ravi2024sam} with 2D bounding boxes near the target objects, leads to segmentation masks of high quality.

We also found that MoGe~\citep{wang2024moge} outputs depths at different scales depending on the input image size. Therefore, we resize all our images to 840x630 for MoGe point cloud annotations.

For 3D bounding box estimation, after filtering the 3D point cloud with a segmentation mask, we run statistical outlier removal and esitmate an oriented 3D bounding box around the remaining points using Open3D~\citep{Zhou2018}.
To facilitate learning, we order all 8 bounding box vertices in a consistent way, starting based on their spatial coordinates with respect to the camera frame.

\subsection{VLA training}
\label{subsec:app-vla-train}

\subsubsection{Data mixture}
\label{subsec:app-vla-data}

We report the VLA training data mixture in \cref{tab:vla-data-mixture}. The sampling weights are chosen manually based on dataset size, visual and task diversity, and quality of language annotations.

\subsubsection{VLA training details}
\label{subsec:app-vla-arch}

\textbf{Reference Frames.}
In this work we focus on learning position control of fixed-base single-arm manipulators.
Each control in the sequence $\mathbf{A_t} = [\mathbf{a}_t, \mathbf{a}_{t+1}, \dots \mathbf{a}_{t+H-1}]$ is defined as a delta with respect to the current end-effector cartesian pose $\Delta_{EE} = [\Delta_{T}, \Delta_{R}]$.
The translation component, $\Delta_{T}$ is in robot base frame and the rotation component, $\Delta_{R}$, is in end-effector frame.
The gripper action is binary.

\textbf{Action Chunking.}
During VLA training we use an action chunk of size $H=5$ and frequency of 5Hz.
As not all datasets in Open X-Embodiment provide action labels at 5Hz, we downsample or upsample the actions accordingly via linear interpolation.
This is done with the goal to encourage the model to share knowledge across datasets with different control frequencies and embodiments instead of 'memorizing' each dataset separately.

\textbf{Architecture.}
Similar to $\pi_0$ \citep{black2024pi_0}, \mname combines a VLM, which processes the image-language inputs, with an \textit{action expert} module, which processes robot proprioception observations and predicts the robot action sequence conditioned on the VLM transformer's intermediate key-value pairs.
The action expert has the same architecture and number of layers as the Gemma \citep{team2024gemma} transformer and configuration downsized to $token\_size=4096, hidden\_size=4096$, for a total of $\sim$300M parameters, which is exactly the same as $\pi_0$~\citep{black2024pi_0}.
Corresponding layers in the VLM and the action expert have a shared attention operation with block-wise causal attention over the blocks $[\mathbf{I}_{t}, \mathbf{l}_t], [\mathbf{p}_t],[\mhat{\mathbf{a}}_{t+1}, \dots, \mhat{\mathbf{a}}_{t+H-1}]$.
Within each block, there is full bidirectional attention and the tokens in each block can attend to tokens in previous blocks, but cannot attend to the tokens in future blocks.
During training, only the action sequence prediction is supervised and gradient updates are propagated back to the VLM parameters through the shared attention layers.

\subsubsection{Flow matching details}
\label{subsec:app-vla-fm}

To address the inherent double coverage of 3D rotations by the unit quaternion group $\mathbb{S}^3$, we ensure that all quaternions used during training and inference lie in the same half-space defined by $\mathfrak{R} (\mathbf{q}) = \mathbf{q}_w > 0 $.

\textbf{Quaternion integration}.
Given a unit quaternion $\mathbf{q}_t \in \mathbb{S}^3$
and its time derivative $ \dot{\mathbf{q}}_t \in \mathbb{R}^4$,
we can compute the angular velocity of rotation via
$\boldsymbol\omega_t = 2.0  \cdot \mathfrak{Im} (\mathbf{q}_t^* \otimes \dot{\mathbf{q}}_t) \in \mathbb{R}^3$.
For a small time step $\Delta t$, the corresponsing delta rotation is given by a rotation vector around the unit axis
$\mhat{\boldsymbol{\omega}} = \boldsymbol{\omega} / ||\boldsymbol{\omega}||$ over an angle $\Delta \phi = ||\boldsymbol{\omega}||\Delta t$.
The corresponding delta quaternion is given by
\begin{equation}
    \Delta \mathbf{q} = \left[ \cos\left(\frac{\Delta \phi}{2}\right), \mhat{\boldsymbol{\omega}} \sin \left(\frac{\Delta \phi}{2}\right) \right].
\end{equation}
The integrated unit quaternion is then given by $\mathbf{q}_{t + \Delta t} = \mathbf{q}_t \otimes \Delta \mathbf{q} \in \mathbb{S}^3$,

\textbf{Rotation losses}.
The denoising vector field for quaternions $\mathbf{u}_t(\mhat{\mathbf{q}}_t^\tau | \mathbf{q}_t) \in \mathbb{R}^4$ is computed as:

\begin{equation}
\begin{aligned}
&\mathbf{u}_t(\mhat{\mathbf{q}}_t^\tau | \mathbf{q}_t)
= \frac{d \mhat{\mathbf{q}}_t^\tau}{d\tau} = \\
&= \frac{\theta}{\sin \theta} \left[
- \cos\!\big((1 - \tau)\theta\big)\,\mathbf{q}_\epsilon
+ \cos\!\big(\tau \theta\big)\,\mathbf{q}_t
\right].
\end{aligned}
\end{equation}

The cosine loss is applied directly on the velocity predictions and has the form:
\begin{equation}
\mathcal{L}_t^{\cos}(\theta) = 1 - \mathbf{v}_\theta(\mhat{\mathbf{A}}^{\tau}_t, \mathbf{o}_t)[\mathbf{q}] \cdot \mathbf{u}(\mhat{\mathbf{A}}^{\tau}_t | \mathbf{A}_t)[\mathbf{q}].
\end{equation}
The geodesic loss is applied on an integrated rotation prediction
$\mhat{\mathbf{q}}_{\theta,t}^{\tau + \delta} \in \mathbb{S}^3$,
derived by integrating the noised input quaternion
$\mhat{\mathbf{q}}_t^\tau$
with the predicted rotation velocity
$\mathbf{v}_\theta(\mhat{\mathbf{A}}^{\tau}_t, \mathbf{o}_t)[\mathbf{q}]$
over a small time step $\delta$.
We follow the integration method described above.
The target is given by the ground truth interpolated quaternion at time $t + \delta$, denoted as $\mhat{\mathbf{q}}_{t}^{\tau + \delta} \in \mathbb{S}^3$.
The closed form geodesic loss is given by:
\begin{equation}
    \mathcal{L}_t^\text{geo} (\theta) = \min |\mhat{\mathbf{q}}_{t}^{\tau + \delta} \pm  \mhat{\mathbf{q}}_{\theta,t}^{\tau + \delta}|.
\end{equation}
The complete rotation loss is given by:
\begin{equation}
    \mathcal{L}_{\mathbb{S}^3} (\theta) = \sum_{k=t}^{t+H} \left[\mathcal{L}_k^{\cos} (\theta) + \mathcal{L}_k^{\text{geo}} (\theta) \right].
\end{equation}

\subsubsection{VLA design decisions details}
\label{subsec:app-vla-design-ablate}

We present more details and results on the design choices we explored for VLA training.

\begin{table}[!ht]
  \centering
  \resizebox{\columnwidth}{!}{
  \begin{tabular}{l|cccc|c}
    \toprule
    Experiment                & \shortstack{Put Carrot\\on Plate} & \shortstack{Put Eggplant\\in Yellow Basket} & \shortstack{Put Spoon\\on Towel} &  \shortstack{Stack Green Block\\on Yellow Block} & \shortstack{Avg. Success\\Rate} \\
    \midrule
    224$\times$224            & \textbf{70.8}\%    & 70.8\%   & 79.1\%    &  8.3\%     & 57.25\% \\
    280$\times$210            & 62.5\% & \textbf{75.0}\% & \textbf{83.3}\% &  \textbf{12.5}\%  & \textbf{58.3}\% \\
    \bottomrule
  \end{tabular}
  }
  \caption{
    \textbf{Image resultion ablations.}
    Different resolutions lead to comparable results on SIMPLER WidowX tasks.
  }
  \vspace{0mm}
  \label{tab:abl_image_res}
\end{table}

\begin{table}[!ht]
  \centering
  \resizebox{\columnwidth}{!}{
  \begin{tabular}{l|cccc|c}
    \toprule
    Experiment                & \shortstack{Put Carrot\\on Plate} & \shortstack{Put Eggplant\\in Yellow Basket} & \shortstack{Put Spoon\\on Towel} &  \shortstack{Stack Green Block\\on Yellow Block} & \shortstack{Avg. Success\\Rate} \\
    \midrule
    trainable SigLIP          & \textbf{75.0}\% & \textbf{100.0}\% & 79.1\% &  \textbf{37.5}\%  & \textbf{72.9}\% \\
    frozen SigLIP             & 62.5\%          & 54.1\%   & 83.3\%    &  25\%     & 56.3\% \\
    frozen-trainable SigLIP   & 66.6\%          & 83.3\%   & \textbf{100.0}\%    &  33.3\%     & 70.8\% \\
    lower lr SigLIP           & 58.3\%          & 58.3\%   & 79.1\%    &  29.1\%    & 56.3\% \\
    \bottomrule
  \end{tabular}
  }
  \caption{
    \textbf{Vision encoder training.}
    Trainable SigLIP outperforms other strategies on SIMPLER WidowX tasks.
    Frozen SigLIP followed by switching on gradients is comparable.
  }
  \vspace{0mm}
  \label{tab:abl_vision}
\end{table}

\begin{table}[!ht]
  \centering
  \resizebox{\columnwidth}{!}{
  \begin{tabular}{l|cccc|c}
    \toprule
    Experiment                & \shortstack{Put Carrot\\on Plate} & \shortstack{Put Eggplant\\in Yellow Basket} & \shortstack{Put Spoon\\on Towel} &  \shortstack{Stack Green Block\\on Yellow Block} & \shortstack{Avg. Success\\Rate} \\
    \midrule
    99-th quantile            & 54.1\%          & 79.1\%   & 79.1\%    &  \textbf{33.3}\%     & 61.5\% \\
    min-max const             & \textbf{66.6}\% & \textbf{87.5}\% & \textbf{87.5}\% &  20.8\%  & \textbf{65.6}\% \\
    mean-std                  & 45.8\%          & 79.1\%   & 45.8\%    &  25.0\%     & 49.0\% \\
    \bottomrule
  \end{tabular}
  }
  \caption{
    \textbf{Translation controls normalization.}
    Normalizing translation controls with min-max constants outperforms other strategies on SIMPLER WidowX tasks.
  }
  \vspace{0mm}
  \label{tab:abl_translation}
\end{table}

\begin{table}[!ht]
  \centering
  \resizebox{\columnwidth}{!}{
  \begin{tabular}{lll|cccc|c}
    \toprule
    \shortstack{Flow\\matching} & \shortstack{Velocity\\Loss} & \shortstack{Rotation\\loss} & \shortstack{Put Carrot\\on Plate} & \shortstack{Put Eggplant\\in Yellow Basket} & \shortstack{Put Spoon\\on Towel} &  \shortstack{Stack Green Block\\on Yellow Block} & \shortstack{Avg. Success\\Rate} \\
    \midrule
    linear & MSE & geodesic & 41.6\%          & \textbf{100.0}\%   & 41.6\%    &  16.6\%     & 50.0\% \\
    linear & cos & geodesic & 41.6\%          & 87.5\%   & 50.0\%    &  29.1\%     & 52.1\% \\
    $\mathbb{S}^3$ & MSE & geodesic & \textbf{45.8}\% & 62.5\% & \textbf{75.0}\% &  \textbf{45.8}\%  & 57.3\% \\
    $\mathbb{S}^3$ & cos & geodesic & \textbf{45.8}\% & 79.1\% & 66.6\% &  \textbf{45.8}\%  & \textbf{59.4}\% \\
    \bottomrule
  \end{tabular}
  }
  \caption{
    \textbf{Linear vs $\mathbb{S}^3$ Flow Matching for rotations.}
    $\mathbb{S}^3$ flow matching consistently outperforms linear flow matching on SIMPLER WidowX tasks.
  }
  \vspace{0mm}
  \label{tab:rot_lerp}
\end{table}

\noindent\textbf{Image Resolution.}
Image resolution ablations are presented in \cref{tab:abl_image_res}.
We observe that square vs 4:3 aspect ratio does not significantly affect performance.

\noindent\textbf{Fine-tuning vision encoders.}
Ablations on fine-tuning vision encoders are presented in \cref{tab:abl_vision}.
Trainable SigLIP strongly outperforms a frozen SigLIP as well as SigLIP with a lower learning rate compared to the rest of the weights.
Freezing SigLIP and fine-tuning for additional 2k steps (frozen-trainable SigLIP) leads to comparable performance to trainable SigLIP, but requires an additional hyperparameter tuning.

\noindent\textbf{Controls normalization.}
Ablations on translation controls normalization are presented in \cref{tab:abl_translation}.
Mean-std normalization is significantly worse than other forms of normalization.
Min-max normalization with const values is slightly better than per-dataset min-max normalization with 1st and 99th quantiles.

\noindent\textbf{Rotations.}
Partial ablations on rotation representations are presented in \cref{tab:rot_lerp}.
$\mathbb{S}^3$ flow matching consistently outperforms linear flow matching.
Cosine distance leads to slightly better performance than MSE for rotation velocity prediction.

\subsubsection{3D ablation details}
\label{subsec:app-vla-3d-ablate}

\textbf{SIMPLER WidowX experiments}. For SIMPLER~\citep{li24simpler} 3D ablations on WidowX, we train on a subset of the Bridge V2~\citep{walke2023bridgedata} dataset, containing demonstrations only from a single kitchen sink environment. The resulting subset comprises $\sim41\%$ of the original Bridge V2 dataset. We train each VLA for 30k steps with batch size 512. Example images from the training and evaluation environments are shown in \cref{fig:3d-widowx}.

\textbf{Franka experiments}. For the 3D ablations on a real-world Franka robot, we train on the entire DROID~\citep{khazatsky2024droid} dataset for 100k steps with batch size 2048. Both models take as input both side and wrist cameras.

\subsection{Real-world robot task description and scoring}
\label{subsec:app-robot-tasks-scoring-rubrics}

We provide the detailed task progression scoring for all real-world evaluations on the WidowX and Franka in \cref{tab:real_world_scoring} and \cref{tab:real_world_scoring_widowx} respectively.

\subsection{Zero-shot control: Bridge V2 vs. DROID}
\label{subsec:app-zero-shot-perf}

\begin{table}[!ht]
    \centering
    \resizebox{\linewidth}{!}{
    \begin{tabular}{|l|c|}
    \hline
        \textbf{Model} & \shortstack{Zero-shot control embodiments\\in real-world \textbf{unseen} environment} \\ \hline
        \textbf{RT-1-X}~\citep{open_x_embodiment_rt_x_2023} & WidowX \\ \hline
        \textbf{RT-2-X}~\citep{open_x_embodiment_rt_x_2023} & WidowX, Google Robot \\ \hline
        \textbf{Octo}~\citep{octo_2023} & WidowX, Google Robot\textbf{?} \\ \hline
        \textbf{OpenVLA}~\citep{kim2024openvla} & WidowX \\ \hline
        \textbf{SpatialVLA}~\citep{qu2025spatialvla} & WidowX \\ \hline
        \textbf{CogACT}~\citep{li2024cogact} & WidowX \\ \hline
        \textbf{FLOWER}~\citep{reuss2025flower} & WidowX \\ \hline
        \textbf{MotoVLA}~\citep{spiridonov2025generalist} & WidowX \\ \hline
        \textbf{CoT-VLA}~\citep{zhao2025cotvla} & WidowX \\ \hline
        \textbf{$\pi_0$}~\citep{black2024pi_0} & \textbf{Franka}, Others\textbf{?} \\ \hline
        \textbf{$\pi_0$-FAST}~\citep{pertsch2025fast} & \textbf{Franka}, Others\textbf{?} \\ \hline
        \textbf{$\pi_{0.5}$}~\citep{intelligence2025pi_} & \shortstack{\textbf{Franka}, Mobile Fibocom,\\Mobile Galaxea, Others\textbf{?}} \\ \hline
        \textbf{Gemini Robotics 1.5} & \shortstack{Bimanual Franka, \\ Aloha, Apollo humanoid} \\ \hline
        \textbf{RDT1} & Bimanual UR5, Aloha \\ \hline
        \textbf{RDT2} & Bimanual UR5, Bimanual Franka \\ \hline
        \textbf{SmolVLA} & S0101 \\ \hline
        \textbf{GROOT-N1.5} & None \\ \hline
        \textbf{SPEAR-1 (ours)} & WidowX, \textbf{Franka}  \\ \hline
    \end{tabular}
    }
    \caption{
        Most works on generalist models for robot manipulation evaluate zero-shot control on Bridge V2 + WidowX using in-distribution environments.
        Only few do so on DROID + Franka in \textbf{unseen} environments.
    }
\end{table} 
Most works on generalist models for robot manipulation \cite{open_x_embodiment_rt_x_2023,kim2024openvla,li2024cogact,zhao2025cotvla,zawalski2024ecot} evaluate the zero-shot control capabilities of their policies by pretraining on the Bridge V2 dataset \cite{walke2023bridgedata} and deploying on the WidowX robot in environments close to the training distribution.
Bridge V2, however, is not very diverse in the number of environments, objects, and camera viewpoints.
As a result, we observe that models pre-trained on Bridge V2 only perform well on WidowX environments when the deployment scenario is similar to what is seen in the dataset (e.g. in the blue toy sink environment), but are usually very sensitive to variations in camera position and OOD backgrounds and objects.
In addition, the WidowX arm has a very low payload and short reach, which makes it unable to manipulate objects beyond the items in a toy kitchen set.
The DROID dataset \cite{khazatsky2024droid}, on the other hand, is significantly more diverse, and the Franka arm used for data collection is more capable.
Furthermore, DROID demonstrations are collected primarily in real-world environments instead of toy environments, the number of unique scenes is $20\times$ higher, and the camera viewpoints vary significantly.
Therefore, we posit that pretraining on DROID and deploying on Franka is a superior experimental setup to showcase generalization to more realistic real-world scenarios, as shown by \cite{atreya2025roboarena}.
The diversity and richness of DROID, however, is at the same time a challenge.
Training generalist control policies on DROID that perform well zero-shot on a Franka robot in unseen environment, is a task that, to the best of our knowledge, has been tackled successfully only by a handful of works so far \cite{pertsch2025fast, intelligence2025pi_}.
In contrast, multiple other works that pre-train on DROID, resort to mixing or fine-tuning on demonstrations collected from the specific target environment in order to achieve good performance~\citep{kim2024openvla,li2024cogact, qu2025spatialvla, reuss2025flower, zhao2025cotvla}.
Therefore, as suggested also by \cite{pertsch2025fast}, we argue that achieving state-of-the-art performance on zero-shot control on the DROID setup by pre-training on DROID is a significantly stronger result than pre-training on Bridge V2 and deploying on WidowX.

\end{document}